\documentclass[runningheads]{llncs}
\usepackage{graphicx}
\usepackage{color}
\usepackage{siunitx}
\usepackage[utf8]{inputenc}
\usepackage[T1]{fontenc}
\usepackage{url}
\usepackage[flushleft]{threeparttable}
\usepackage{multirow}
\usepackage{color}
\usepackage{amsmath}
\usepackage{caption}
\usepackage{subcaption}
\usepackage{rotating}
\newenvironment{item2}
{\begin{list}{$\bullet$}{
\leftmargin=6mm
\labelwidth=3mm
\labelsep=3mm
}}{\end{list}}
\begin{document}
%
\title{Decoding machine learning benchmarks}
\titlerunning{Decoding machine learning benchmarks}
%
\author{Lucas F. F. Cardoso\inst{1}\orcidID{0000-0003-3838-3214}\and
Vitor C. A. Santos \inst{1}\orcidID{0000-0002-7960-3079}\and
Regiane S. Kawasaki Francês\inst{1}\orcidID{0000-0003-3958-064X}\and
Ricardo B. C. Prudêncio\inst{2}\orcidID{0000-0001-7084-1233} \and
Ronnie C. O. Alves\inst{3}\orcidID{0000-0003-4139-0562}}

\authorrunning{Cardoso, L. et al.}
%
\institute{Faculdade de Computação, Universidade Federal do Pará, Belém, Brazil\\
\email{lucas.cardoso@icen.ufpa.br, vitor.cirilo3@gmail.com, kawasaki@ufpa.br}\and
Centro de Informática, Universidade Federal de Pernambuco, Recife, Brazil\\
\email{rbcp@cin.ufpe.br} \and
Instituto Tecnológico Vale, Belém, Brazil\\
\email{ronnie.alves@itv.org}}
%
\maketitle              
\begin{abstract}
Despite the availability of benchmark machine learning (ML) repositories (e.g., UCI, OpenML), there is no standard evaluation strategy yet capable of pointing out which is the best set of datasets to serve as gold standard to test different ML algorithms. In recent studies, Item Response Theory (IRT) has emerged as a new approach to elucidate what should be a good ML benchmark. This work applied IRT to explore the well-known OpenML-CC18 benchmark to identify how suitable it is on the evaluation of classifiers. Several classifiers ranging from classical to ensembles ones were evaluated using IRT models, which could simultaneously estimate dataset difficulty and classifiers' ability. The Glicko-2 rating system was applied on the top of IRT to summarize the innate ability and aptitude of classifiers. It was observed that not all datasets from OpenML-CC18 are really useful to evaluate classifiers. Most datasets evaluated in this work (84\%) contain easy instances in general (e.g., around 10\% of difficult instances only). Also, 80\% of the instances in half of this benchmark are very discriminating ones, which can be of great use for pairwise algorithm comparison, but not useful to push classifiers abilities. This paper presents this new evaluation methodology based on IRT as well as the tool decodIRT, developed to guide IRT estimation over ML benchmarks.

\end{abstract}

\section{Introduction}

Machine learning (ML) is a field  of artificial intelligence which has been rapidly growing in recent years, partly due to the diversity of applications in different areas of knowledge. 
There are different types of machine learning algorithms, from unsupervised to supervised ones \cite{b4}. In this work, we focus on supervised algorithms, and more specifically on  classification algorithms, which are commonly adopted in many applications \cite{b1}.


Empirically evaluating ML algorithms is crucial for assessing the advantages and limitations of the available techniques. Algorithm evaluation is usually performed by deploying datasets available in repositories  \cite{b2}. Remarkably, the OpenML repository has been developed as an online platform, allowing researchers to share results, ML strategies and datasets used in benchmarking experiments. Such platform improves reproducibility and minimizes double effort to do the same experiment \cite{b2}. Another contribution is the OpenML Curated Classification (OpenML-CC18), a benchmark for classification that has 72 standardized and cured datasets \cite{b8}.

By relying on benchmark datasets, ML models can be trained and tested using a chosen experimental methodology (e.g., cross-validation) and the performance measures of interest (e.g., success rate). For increasing robustness, different datasets can be adopted in the experiments and the performance measures are  averaged across the datasets. This approach however does not allow a deeper analysis of classifier's capacity, since the type of dataset adopted in an experiment may bias the obtained results. For instance, the performance of a ML algorithm can be overestimated if only easy datasets are adopted in the experiments. Also, depending on the datasets, certain ML algorithms may be favored, thus giving the false impression that one classifier is the best in relation to the others \cite{b5}.


Given the above context, it is important that the performance analysis of a classifier takes into account the complexity of the datasets as well. Previous work \cite{b6,b5,b7} has addressed this issue by adopting concepts from Item Response Theory (IRT) to provide a more robust approach to devise benchmarks. IRT is commonly used in psychometric tests to assess the performance of individuals on a set of items (e.g., questions) with different levels of difficulty. IRT has been extended to ML evaluation by treating classifiers as individuals and test instances as items. Thus ML algorithms can be ranked by IRT according to their ability to correctly respond the most difficult instances.


The main contributions of this work are: I) an empirical evaluation of IRT to evaluate ML benchmarks, adopting the OpenML-CC18 benchmark as a case study; II) a global assessment of ML algorithms in benchmarks comparison, improving the identification of the strengths and weaknesses among them on the basis of IRT estimators; and III) the decodIRT tool, developed to guide the IRT estimation over ML benchmarks automatically along with a robust rating system to establish proper boundaries between ML algorithms. Regarding the third contribution, rating systems are widely used in other tasks (e.g., chess playing) to indicate the strength of an individual in a competition \cite{b9}. Among the existing systems, the Glicko-2 \cite{b10} system was used in order to create a ranking that is able to summarize the algorithms' results generated by the IRT. Thus it pointed out the most suitable ML algorithms for the classification tasks proposed by the OpenML-CC18 benchmark.



The remaining of this paper is organized as follows: Section 2 contextualizes the main subjects explored in this work, specifically w.r.t. OpenML, Item Response Theory and the Glicko-2 system. Section 3 presents the methodology used, explains how decodIRT tool works and how the Glicko-2 system can be applied to summarize IRT results. Section 4 presents an overall discussion. Section 5 concludes the article and future developments.

\section{Background}

\subsection{OpenML}

OpenML is a repository in which machine learning researchers can share data sets, descriptions of experiments and obtained results in as much detail as possible. Hence, a better organization and use of such information is allowed, thus creating a collaborative environment for sharing experiments in a global scale \cite{b2}. In OpenML anyone can download a dataset, execute a machine learning method of their choice and share the results obtained with other users, generating discussions and new information about the data sets and the applied algorithms.


In addition, the platform also provides several sets of reference datasets. OpenML-CC18 \footnote{Link to access OpenML-CC18: \url{https://www.openml.org/s/99}} is one of those reference sets that includes 72 existing datasets in OpenML from mid-2018 and that meet several requirements in order to compile a complete reference set. OpenML-CC18 also includes datasets frequently used in benchmarks published in recent years \cite{b8}. Several associated metadata is available such as: number of instances, number of features, number of classes and proportion between the minority and majority class of the group.

OpenML actually has a lot to contribute to research in the field of machine learning. The current work aims to apply IRT to evaluate this gold standard to provide greater strength for OpenML-CC18 and ML community. The analysis of the datasets within the IRT perspective allows adding to the reference set new relevant metadata, such as the difficulty of the datasets and discriminatory potential of the data. Such information can be very useful for choosing a benchmarking set and will be better discussed in the following sections. 

\subsection{Item Response Theory}

Andrade, Tavares and Valle (2000) \cite{b11} describe IRT as a set of mathematical models that aim to represent the probability that an individual will correctly answer an item according to the item's parameters and the respondent's ability, so that the greater the ability, the greater the chance of a correct response.

The item can be classified according to its associated response. It can be dichotomous, when the responses are just right or wrong, or non-dichotomous, if there are more than two possible answers. Logistical models for dichotomous items are the most used in literature and practice. There are basically three types of models, which differ by the number of item parameters that are estimated. In the 3-parameter logistic model, known as 3PL, the probability of an individual $j$ correctly respond an item $i$ given his ability is defined by the following equation:

\begin{equation}
    P(U_{ij} = 1|\theta_{j}) = c_{i} + (1 - c_{i})\frac{1}{1+ e^{-a_{i}(\theta_{j}-b_{i})}}
\end{equation}

\noindent Where:

\begin{item2}
\setlength\itemsep{.25cm}
  
  \item $U_ {ij}$ is the dichotomous response that can take the values 1 or 0, being 1 when the individual \textit{j} hits the item \textit{i} and 0 when he misses;
  
  \item $\theta_{j}$ is the ability of the individual \textit{j};
  
  
  \item $b_{i}$ is the item's difficulty parameter and indicates the location of the logistic curve;
  
  \item $a_{i}$ is the item's discrimination parameter, i.e., how much the item \textit{i} differentiates between good and bad respondents. This parameter indicates the slope of the logistic curve. The higher its value, the more discriminating the item is;
  
  \item$c_{i}$ is the guessing parameter, representing the probability of a casual hit. It is the probability that a respondent with low ability hits the item.

\end{item2}

The 2PL can be derived by simplifying the above model and dropping the guessing parameter, i.e., $c_{i}=0$. Finally, the 1PL parameter can be defined dropping the discrimination parameter, i.e., assuming $a_{i} =1$. 

Although it is possible to obtain negative values of discrimination, they are not expected by IRT. The negative values are not expected because it means that the probability of a hit is higher for individuals with lower ability values, rather than higher than what is normally expected \cite{b11}.

The logistic curves are estimated from responses collected for a group of items and respondents. According to Martínez-Plumed et al (2016) \cite{b5}, there are three possible situations for estimation. In the first situation, the parameters of the items are known, but the ability of the respondents is not known. The second possible situation is just the opposite: the ability of each respondent is defined, but the parameters of the items are not known. In the third case, which is also the most common, both the parameters of the items and the abilities of the respondents are unknown. This work is in the third case and for this situation, the following two-step interactive method proposed by Birnbaum is used \cite{b15} for estimation from the group of responses:

\begin{itemize}
  \item At first, the parameters of each item are calculated with only the answers of each individual. The initial values of respondent's ability can be the number of correct answers obtained. 
  In the case of classifiers, this work used the obtained accuracy as the initial ability;
  \item Once the parameters of the items are obtained, the ability of the individuals can be estimated. For both item's parameters and respondent's ability, simple estimation techniques can be used, such as the Maximum Likelihood Estimate \cite{b5}.
\end{itemize}


IRT is usually applied for educational purposes, in which respondents are students and items are questions. Recently, IRT has been extended to AI, and more specifically to ML, in which respondents are assumed to be AI techniques and items are AI tasks \cite{b6,b5,b7}. To analyze datasets and learning algorithms through IRT. In this work, it was used the instances of a dataset as the items and the classifiers were assumed as the respondents. Among the existing IRT models, the 3PL logistic model was used because it is the most complete. The IRT parameters were used to assess the datasets directly, informing the percentage of instances with high difficulty, with great discriminative power and with high chance of casual hit. This allows to have an insight into the complexity of the datasets evaluated and how different classifiers behave in the face of the challenge of classifying different datasets.

\subsection{Glicko-2 System}

The Glicko-2 system is an extension of Glicko, originally developed by Mark E. Glickman \cite{b10} to measure the ``strength'' of chess players. Today, the system is used worldwide by several organizations such as the Australian Chess Federation \cite{b13}.


In the Glicko-2 system, each player has three variables used to measure his statistical strength: the R rating, the RD rating deviation and the $\sigma$ volatility. The R rating is the numerical value itself that measures the strength of a player. As it cannot be said that this value accurately represents the strength of an individual, the RD is used to indicate the reliability of an R rating. The Glicko-2 system estimates the players' strength in a 95\% confidence interval, as follows: $[R-2RD, R + 2RD]$. Thus, there is a 95\% chance of the real player's strength is within the calculated range. The lower a player's RD, the more accurate his rating value is. In addition, the $\sigma$ volatility makes it possible to measure the degree of expected rating fluctuation, i.e., the higher the volatility value, the greater the chances of a player's rating fluctuating within his RD range. Therefore, the lower the value of $\sigma$, the more reliable and expected a player's rating is. For example, among players with low volatility, it is possible to state more precisely who is the strongest based on their ratings \cite{b13}


To estimate the rating values of the players, the Glicko-2 system uses the so-called classification period. This period is a sequence of games played by the individual in question. At the end of this sequence, the Glicko-2 system can update the player's parameters. To do this, the Glicko-2 system uses the opponents' rating and RD values, along with the score of the game result (e.g., 0 points for defeat and 1 points for victory). If there is no previous player data, the Glicko system uses default initial values for new players, 1500 for rating, 350 for RD and 0.06 for $\sigma$ \cite{b10}.

\section{Materials and methods}

In order to build the IRT models and analyze ML benchmarks, it was developed a tool called decodIRT \footnote {Link to the source code: \url{https://github.com/LucasFerraroCardoso/IRT_OpenML}}. The tool was written mostly in Python together with R language features. The decodIRT has the main objective of assisting the analysis of existing datasets on the OpenML platform as well as the proficiency of different classifiers. For this, it relies on the probability of correctness derived by the IRT model as well as the estimated items' parameters and respondents' ability. 

\begin{figure}[htbp]
\centerline{\includegraphics[width=.8 \textwidth]{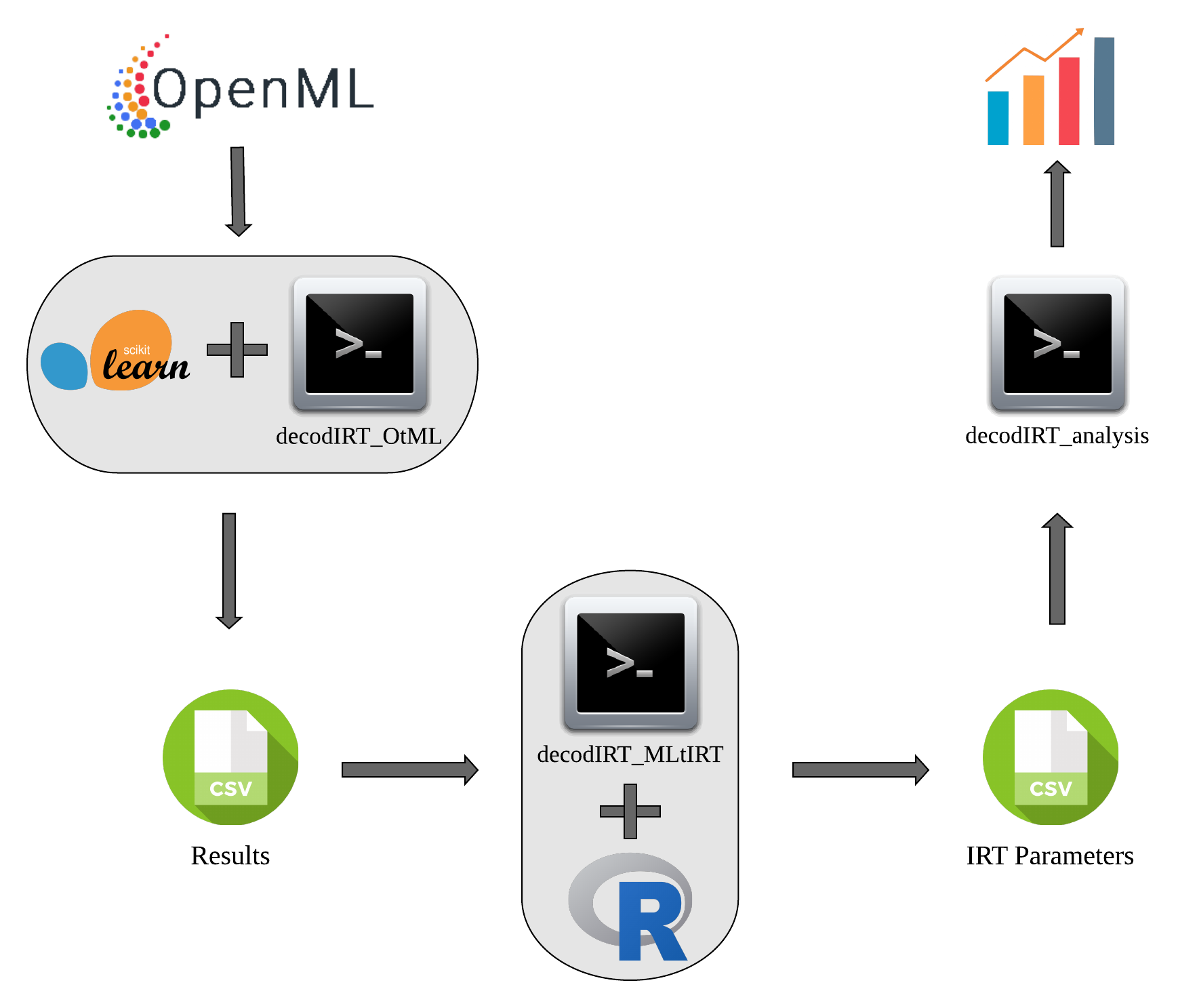}}
\caption{Flowchart of the decodIRT execution.}
\label{fig1}
\end{figure}

As it can be seen in Figure~\ref{fig1}, the decodIRT tool consists of three scripts designed to be used in sequence. The first script is in charge of downloading the OpenML datasets and generating several ML models and putting them to classify the datasets. It will generate a response matrix, which contains the result of the classification of each test instance. The response matrix will be the input for the second script, which in turn is in charge of calculating the item parameters. The last script will use the data generated by the previous scripts to rank the datasets using the item parameters and use the IRT to analyze the ML models.


\subsection{decodIRT\_OtML}

 The first script will download the existing OpenML datasets and execute several classifiers implemented in the Scikit-learn \cite{b20} package. These classifiers will generate the response data that will be used to estimate the item parameters. After downloading the dataset, the stratified split for training and testing is performed, to maintain the proportions of the classes. By default, the 70/30 split is used. However, for very large datsets the split is adapted to leave a maximum of 500 instances for testing. Martínez-Plumed et al. (2019) \cite{b7} explain that for a very large number of items, packages that estimate IRT values may be stuck at the local minimum or may not converge. Therefore, for very large datasets it is recommended to use less than a thousand instances to generate the parameters.

In order to generate diverse response values, at first, 120 models of Neural Networks (MLP) are executed, varying only the depth of the networks in which the depth starts at 1 and gradually increases to 120. Each of the MLPs is trained using the 10-fold cross-validation.

After that, classifications are performed with a second set of classifiers that will be evaluated by the tool, which are: Naive Bayes Gaussian standard, Naive Bayes Bernoulli standard, KNN of 2 neighbors, KNN of 3 neighbors, KNN of 5 neighbors, KNN of 8 neighbors , Standard Decision Trees, Random Forests (RF) with 3 trees, Random Forests with 5 trees, Standard Random Forests, Standard SVM and Standard MLP. The models defined with "default" means that the default parameter values that Scikit-learn provide has been used for each ML algorithm.

Based on the work of Martínez-Plumed et al. (2016) \cite{b5}, It was inserted 7 artificial classifiers, three random (classify randomly), a majority classifier (classifies all instances with the predominant class), a minority classifier (classifies all instances with the non-predominant class), a very bad one (it misses all classifications) and a great one (it gets all classifications right). Artificial classifiers serve to set proper performance indicators regarding ideal classification boundaries.

\subsection{decodIRT\_MLtIRT}

In the second script, the IRT item parameters are calculated based on the classifier responses that were generated in the previous step. For this, it was necessary to communicate with the R language through the use of Rpy2 library. The R language was used due to the Ltm package, which implements a framework containing several mechanisms for the calculation and analysis of IRT for dichotomous data, as well as the generation of item parameters.

As mentioned previously, a maximum of 500 instances were used to generate the item parameters. This limit was chosen since for larger values it was not possible to generate the parameters for some datasets.

\subsection{decodIRT\_analysis}

The last step is in charge of doing the analysis on the data generated by the previous steps. By definition, the code will generate in all executions a ranking of the datasets on the percentage of test instances, whose parameter value exceeds the defined limit. For example, 50\% of instances have a difficulty value above 1.

If the user does not define any limit for the parameters, standard limit values based on Adedoyin, Mokobi et al. (2013) \cite{b16} are used. In Adedoyin's work it is said that to consider an item (instance) as being difficult its value of Difficulty should normally be higher than 1. Items with high discriminative capacity should normally have Discrimination values greater than 0.75, and items with high guessing values are normally greater than 0.2. The analysis of the percentages of each item parameter is one of the interests of this work.

Following the Birnbaum method, to calculate the probability of success for each classifier, the proficiency of each classifier is estimated first. To do so, Python's Catsim package is used, which implements both functions to calculate the probability of success, as well as functions to estimate proficiency using the responses of each classifier and the item parameters.

In order to analyze more generally how the different classifiers perform on the datasets. The concept of True-Score \cite{b17} was also implemented. The True-Score is the sum of all the probabilities of a student's (classifier) correct answer to the questions of a test. The purpose of using True-Score is to calculate a final grade, just as we get when we take a test. Thus, it is expected to identify the "aptitude" that the different classifiers have before being presented a dataset. A conceptual view of the innate ability, engineering of the model, through its default hyperparameters settings, it will be put to the test by the Glicko-2 system.

\subsection{Ranking of classifiers by the Glicko-2 system}
Glicko-2 \cite{b10} has been used to estimate the strength of classifiers and to generate a global performance assessment by simulating a competition among all ML methods. The simulation explores a round-robin tournament, thus all classifiers face each other in the championship and at the end there is a ranking that will reflect how each ML algorithm performed over the competition.

Each dataset is seen as the championship phases and is used as a classification period, so the classifiers face each other in each dataset. The True-Score values obtained by each classifier are used to decide the winner of the match and this is done as follows: if the score is higher than that of your opponent, it is counted as victory, if it is lower it is counted as defeat and if it is the same it is given as a tie. For that, the scoring system used in chess competitions is used, with 1 point for victory, 0 for defeat and 0.5 for a tie. Thus, at the end of each dataset, all classifiers will have their rating, RD and volatility values. These new values are used as the initial values for the next dataset. Once all datasets are finalized, the rating values generated are used to create the ranking of classifiers.

\section{Results and discussion}

One of the main objectives of this work is to use the IRT item parameters to give more reliability to the use of the OpenML-CC18 benchmark. The item parameters explored are those of difficulty and discrimination. Although there are 72 datasets available in OpenML-CC18, only 60 were evaluated by decodIRT. This is due to two reasons: (1) the size of datasets, being 11 too large and require a long time for the execution of all ML models; (2) It was not possible to generate the item parameters for the “Pc4” dataset. The latter is still an open issue. Thus, the OpenML-CC18 benchmark evaluated in this work corresponds to 83.34 \% of the total reference set. These datasets are then used to performance analysis \footnote{All classification results can be obtained at \url{https://github.com/LucasFerraroCardoso/IRT_OpenML/tree/master/benchmarking}}.

\subsection{Decoding OpenML-CC18 Benchmark}

\begin{figure}[htbp]
\centerline{\includegraphics[width=.7 \textwidth]{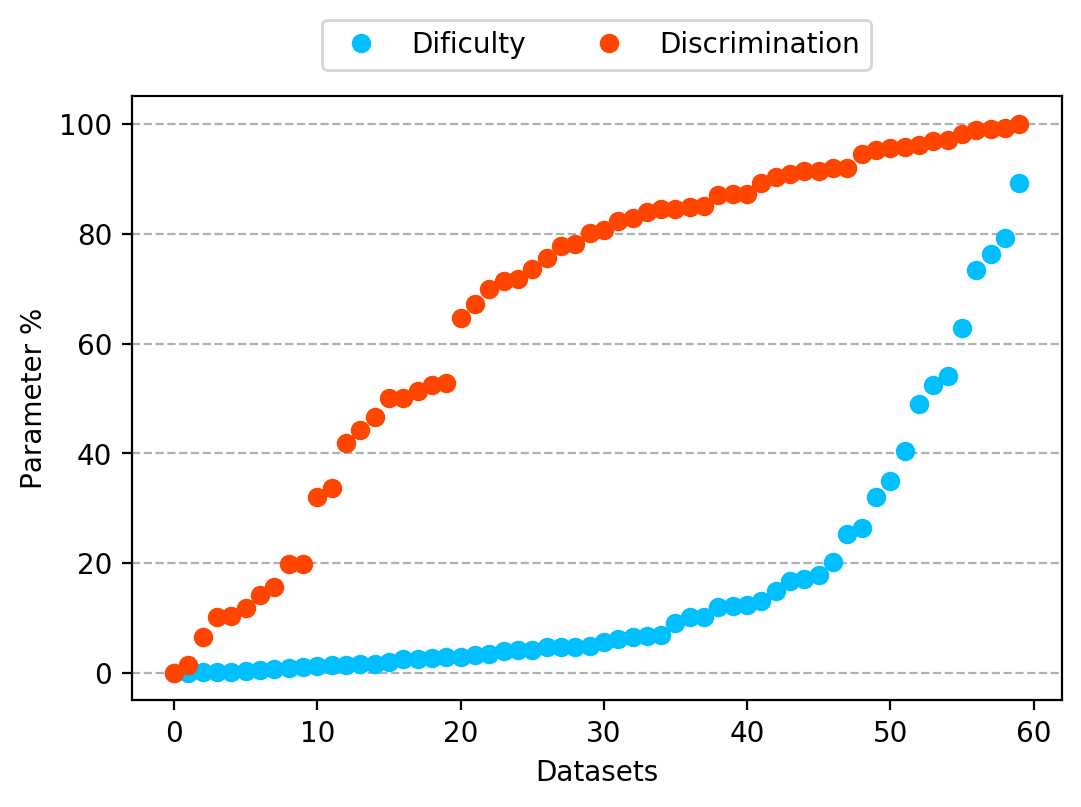}}
\caption{It shows the percentages of difficult and very discriminative instances arranged in ascending order. There is a certain percentage of discrimination and a percentage of difficulty that are in the same position on the X axis do not necessarily correspond to the same dataset. ``tic-tac-toe'', ``credit-approval'' and ``optdigits'' are respectively the datsets with the most difficult instances. While ``banknote-authentication'', ``analcatdata\_authorship'' and ``texture'' are the most discriminative.}
\label{fig2}
\end{figure}


It was possible to observe that there is a inverse relationship between difficulty and discrimination parameters. The rankings highlights that the most difficult datasets are also the least discriminative and vice versa (Figure~\ref{fig2}). Thus, the difficulty posed by these datasets, despite make them more challenging, they are less suitable to differentiate classifiers. On the other hand, the most discriminatory datasets are not suitable for testing classifiers abilities.

Among the analyzed datasets, 49 have less than 27\% of instances considered difficult, while only 7 have more than 50\% difficult. This means that of the 60 datasets observed, 81.67\% of the total have more than 70\% of their instances considered easy and only 11.67\% are difficult. Therefore, the OpenML-CC18 benchmark, for model comparison purposes, should be adopted with caution. Figure~\ref{fig2} illustrates graphically the classification challenge presented in this benchmark. It is clear that the situation is reversed when compared to the difficulty (blue line) values. In which only 25\% of datasets have less than 50\% of their instances considered to be less discriminatory and 31 of the 60 datasets have at least 80\% of very discriminative instances. From this observation one can infer that although the OpenML-CC18 is not considered as difficult as it was expected, it does have appropriate datasets to differentiate good and bad classifiers. In addition, knowing which datasets are more difficult, allows the user to choose his benchmark set more specifically, in case he wants to test the classification power of his algorithm, disregarding the need to test with all datasets. In summary, focusing on datasets that present instances with a higher positive discrimination values.

\subsection{Classifiers performance on OpenML-CC18}

\begin{figure}[htbp]
\centerline{\includegraphics[width=1 \textwidth]{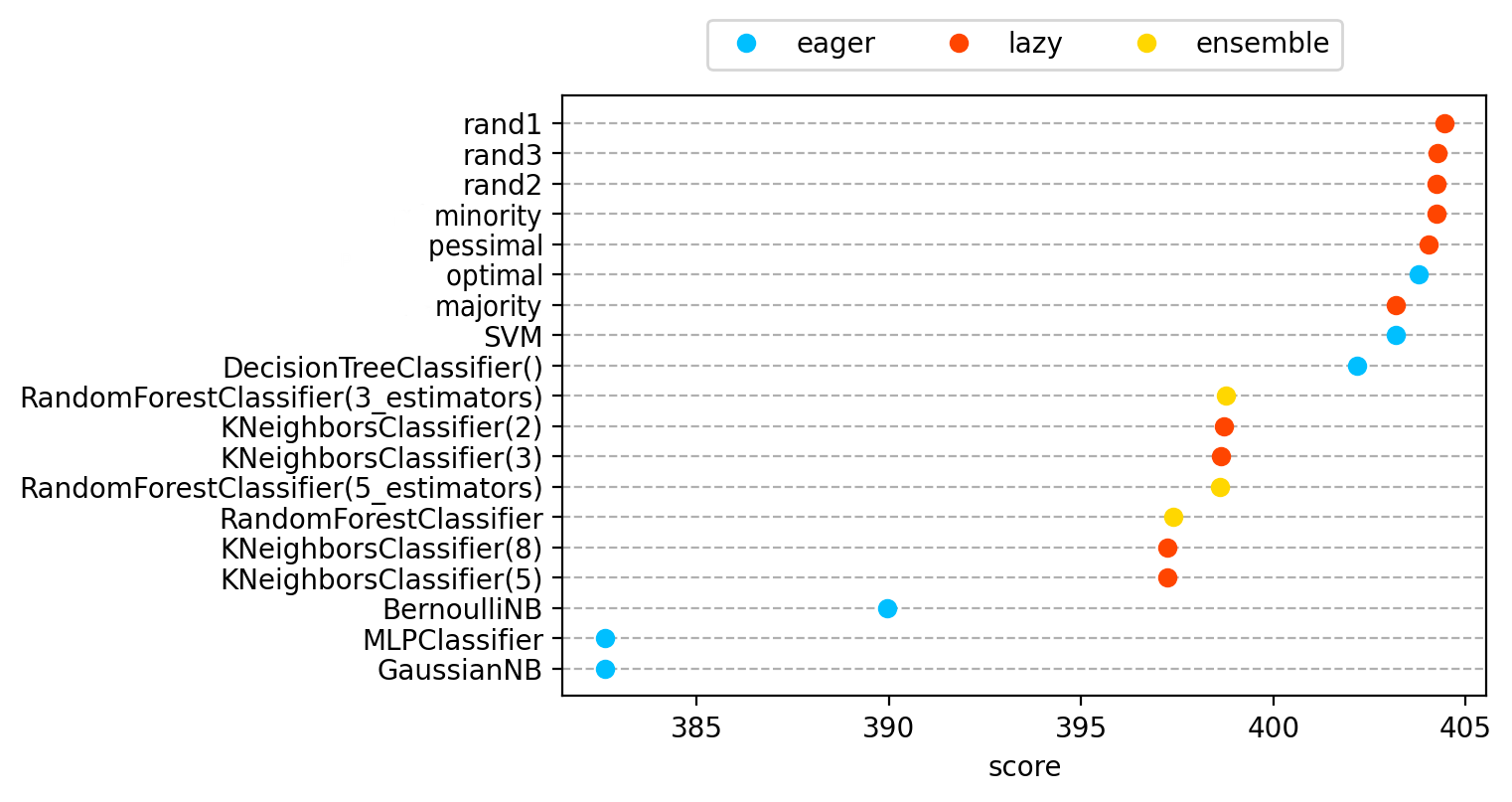}}
\caption{The True-Score values obtained for the ``jm1'' dataset.}
\label{fig3}
\end{figure}

Taking into account only the values of True-Score \footnote{All data generated can be accessed at \url{https://github.com/LucasFerraroCardoso/IRT_OpenML/tree/master/BRACIS}} of the classifiers for each dataset, it possible to see a pattern in the overall performance of the classifiers. As expected, the optimal classifier has the highest score and the other artificial classifiers, the worst results. In addition, there are cases where the real classifiers match the score of the optimum and even surpass it. However, in some specific cases the score is reversed and the best classifiers have the worst scores (see Figure~\ref{fig3}).

The score of the optimal classifier be surpassed and the last case mentioned above can happen due to the occurrence of many instances having negative values of discrimination. Therefore, it can be inferred that datasets having many negative instances may not be good for creating benchmarks. A future work would be to carefully analyze how the characteristics of a dataset can cause such situations, in addition to observing how these characteristics can positively or negatively influence the performance of the classifiers. OpenML already has an extensive set of metadata which can be used for this purpose.

Although the True-Score values generated already allow observing the performance of the classifiers, there is still a lot of data to be evaluated individually, towards a natural assessment of the innate ability of these learning algorithms. Thus, the Glicko-2 rating system was used to summarize this information. Table 1 shows the final rating ranking that was obtained.

\begin{table}[htbp]
\caption{Classifier rating ranking}
\begin{center}
\begin{tabular}{|c|c|c|c|}
\hline
\textbf{\textit{Classifier}} & \textbf{\textit{  Rating  }} & \textbf{\textit{  RD  }} & \textbf{\textit{ Volatility }}\\
\hline
optimal                                 & 1732.56 & 33.25 & 0.0603     \\ 
MLPClassifier                         & 1718.65 & 31.20 & 0.0617     \\ 
RandomForestClassifier                & 1626.60 & 30.33 & 0.0606     \\ 
RandomForestClassifier(5\_estimators) & 1606.69 & 30.16 & 0.0621     \\ 
RandomForestClassifier(3\_estimators) & 1575.26 & 30.41 & 0.0646     \\ 
DecisionTreeClassifier()              & 1571.46 & 31.16 & 0.0674     \\ 
SVM                                   & 1569.48 & 32.76 & 0.0772     \\ 
KNeighborsClassifier(3)               & 1554.15 & 30.74 & 0.0646     \\ 
GaussianNB                            & 1530.86 & 31.25 & 0.0683     \\ 
KNeighborsClassifier(2)               & 1528.41 & 30.40 & 0.0638     \\ 
KNeighborsClassifier(5)               & 1526.10 & 30.27 & 0.0630     \\ 
BernoulliNB                           & 1494.87 & 32.64 & 0.0770     \\ 
KNeighborsClassifier(8)               & 1457.78 & 30.25 & 0.0638     \\ 
minority                           & 1423.01 & 30.66 & 0.0631     \\ 
rand2                                 & 1374.78 & 30.27 & 0.0605     \\ 
rand3                                 & 1337.27 & 30.95 & 0.0600     \\ 
rand1                                 & 1326.38 & 31.42 & 0.0610     \\ 
majority                           & 1301.08 & 31.74 & 0.0666     \\ 
pessimal                               & 1270.46 & 31.74 & 0.0603     \\ \hline
\end{tabular}
\label{tab1}
\end{center}
\end{table}


The positions of the artificial classifiers in the ranking were as expected. The excellent and very bad classifiers assumed the first and last positions, respectively, while the other artificial classifiers performed less than all real classifiers. Among the real classifiers, MLP was the winner, being slightly better than RF classifiers.

When looking at the position of the real classifiers, it is surprising how the MLP got very close to the rating of the optimal classifier, while the other classifiers were more than 100 points below the first place. MLP's ``strength'' can also be confirmed by looking at the volatility values of the classifiers, in which all have low values. Such values also help to give more confidence with respect to the position of each classifier in the ranking and allow us to infer that the rating values in fact represent the ``strength'' of the classifiers with precision. Considering the rating fluctuation of each classifier within its respective RD range, the final positions may change. The MLP, for example, if it had its ``strength'' at the lowest value by your RD range, would have a new rating of 1656.25. This would allow classifiers who are 3rd and 4th to overtake if their rating values fluctuate upwards. However, from the 5th position, no classifier can reach even if its rating fluctuates to the maximum of its range.

It can be observed that there are groups of classifiers that have equivalent ``strength'', in which it is not possible to define, with high confidence, which one of the them present highest aptitude in OpenML-CC18 challenge. This ``strength'' is translated into the aptitude of these learning algorithms. Since tests were performed with several datasets presenting distinct IRT estimators, and classifiers have same configuration model settings, it is assumed that the all classification results indeed reflects the learning algorithm aptitude, by design. It is important to note that optimization can have an effect of fine-tuning the decision boundaries in the most difficult datasets, and consequently a better performance. On the other hand, it would not allow the assessment of innate, engineering skills of these classifiers. A future work could stress how far one can go in ML optimization, extrapolating the boundaries between ability and aptitude.

\begin{figure}[htbp]
\centerline{\includegraphics[width=0.85 \textwidth]{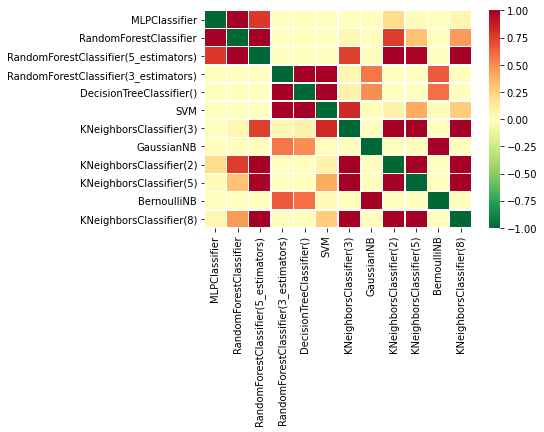}}
\caption{Heatmap generated from the Nemenyi Test, using only the rating distributions of the real classifiers.}
\label{fig4}
\end{figure}

In order to bring more credibility to the generated rating values, the Friedman \cite{b18} test was carried out. The objective is to identify whether the rating values, in fact, allow to differentiate the ``innate ability'' of the classifiers. The Friedman test was performed using the rating values of the real classifiers. The execution of the Friedman Test resulted in a p-value of approximately \num{9.36e-80}. Subsequently, the Nemenyi \cite{b19} test was applied to compare and identify which of the distributions differ from each other. Figure~\ref{fig4} presents a Heatmap of the Nemenyi test.

When analyzing the Heatmap, one can observe that the three classifiers having the highest ratings have also high p values, so they do not differ from each other. And even though they are the more performant, they all have high values for at least one different classifier. The other classifiers also have performance similarities, even though they are from different families in some cases. Therefore, it is not evident that there is a clear separation of classifiers into different groups, since some classifiers that are at the top of the ranking do not differ from all classifiers that are in lower positions. This leads us to believe that, although the Friedman Test points out that there are different ML groups, these differences are not statistically significant enough to identify the highest skilled one.

\section{Final considerations}

This work explores IRT estimation for the evaluation ML benchmarks. ML benchmarks are commonly used to explore how far ML algorithms can go while handling datasets. The OpenML-CC18 is a gold standard. Although the ML community make broad use of it, such utilization might be taken with caution. From the 60 datasets evaluated in this work, only 12\% of their instances are considered difficult; 80\% of the instances in half of this benchmark are very discriminating ones which can be of great source for comparisons analysis, but not useful to push classifiers abilities. The benchmark assessment methodology is provided and it can be reproduced using the decodIRT tool. Even though classifier abilities are highlighted by IRT, there were also an issue regarding the innate ability, aptitude, whether one can set the borders between the ML algorithm (by design) and training (optimization). The IRT results were explored by rating systems such as the ones used to evaluate the strength of chess players to establish the ML winner, and consequently providing a glimpse towards a aptitude score of ML algorithms.

\end{document}